\documentclass[letterpaper, 10 pt, conference]{ieeeconf}  

\usepackage{hyperref}
\usepackage{amsmath} 
\usepackage{amssymb}
\usepackage{graphicx} 
\usepackage{acronyms}
\usepackage{mfirstuc}
\usepackage{amsfonts}
\usepackage{xcolor} 
\usepackage{comment}
\usepackage{afterpage}
\usepackage{multirow}
\usepackage{comment}
\usepackage{algorithm}
\usepackage{siunitx}
\usepackage{algpseudocode}
\usepackage{tikz}
\usetikzlibrary{positioning}
\usepackage{pgfplots}
\pgfplotsset{compat=1.17}

\usepackage{float}
\usepackage[figurename=Fig.]{caption}
\usepackage{booktabs}
\usepackage{tabularray}
\usepackage{svg}
\usepackage{subcaption}
\usepackage[symbol]{footmisc} 
\usepackage[noadjust]{cite}

\renewcommand{\thefootnote}{\fnsymbol{footnote}}

\IEEEoverridecommandlockouts                              

\title{\LARGE \bf
\capitalisewords{In Situ Monitoring and Predictive Modeling of Harmful Algal Blooms with Autonomous Surface Vehicles}}

\author{Nenyi Kweku Nkensen Dadson$^{1}$, Jean Paul Gonzalez$^{2}$, Daniel P. Arabie$^{2}$, Ryan J. Cedel$^{1}$,  \\ Dorin Boldor$^{3}$, Corina Barbalata$^{1}$
\thanks{$^{1}$Department of Mechanical \& Industrial Engineering,   Louisiana State University, Baton Rouge, LA 70803, USA; $^{2}$Electrical and  Computer Engineering Division, Louisiana State University, Baton Rouge, LA 70803, USA; $^{3}$Louisiana State University Agricultural Center, Louisiana State University, Baton Rouge, LA 70803, USA
        {\tt\small ndadso1@lsu.edu; darabi8@lsu.edu; jgonz95@lsu.edu; rcedel1@lsu.edu; DBoldor@agcenter.lsu.edu; cbarbalata@lsu.edu}}%
}

\begin{document}

\maketitle
\thispagestyle{empty}
\pagestyle{empty}
\begingroup
\renewcommand{\thefootnote}{}
\footnotetext{%
This paper has been accepted for publication in
the Proceedings of OCEANS 2026. This is the author-accepted
manuscript; the final published version will be available
through IEEE Xplore.

\par\smallskip

\copyright~2026 IEEE. Personal use of this material is permitted.
Permission from IEEE must be obtained for all other uses,
in any current or future media, including reprinting/republishing
this material for advertising or promotional purposes,
creating new collective works, for resale or redistribution
to servers or lists, or reuse of any copyrighted component
of this work in other works.
}
\endgroup

\begin{abstract}
\acfp{HAB} pose significant environmental, economic, and public-health challenges in coastal and inland water bodies, creating a need for scalable methods to collect spatially distributed water-quality data. This paper presents an autonomous surface vessel-based sensing system for in-situ water-quality monitoring related to harmful algal bloom detection. The system integrates a BlueBoat \ac{ASV}, a motorized winch for depth-controlled sonde deployment, a multiparameter water-quality sonde, side-scan sonar, and an event-camera payload. Autonomous waypoint missions were used to collect measurements across selected sampling regions and multiple depths, enabling both horizontal and vertical characterization of environmental conditions. Field deployments were conducted in Lake Pontchartrain, Louisiana, with additional measurements collected in False River, Louisiana to increase dataset diversity. The collected measurements included algal pigment indicators, nutrient concentration, turbidity, temperature, and other water-quality parameters relevant to bloom formation. In addition, complementary sonar and event-camera data were collected during vehicle operation to provide contextual environmental information.  A machine-learning classification approach was also evaluated using the collected water-quality measurements to support \ac{HAB} and non-\ac{HAB} classification. The results demonstrate the feasibility of using a modular \ac{ASV} platform for repeatable, multi-modal water-quality data collection and support future development of autonomous monitoring systems.
\end{abstract}

\vspace{-10pt}
\section{Introduction} 
\label{sec:introduction}
Algae are often considered as one of the most important plant species on earth; contributing to about $30\%$ to $50\%$ of the available oxygen in the ecosystem \cite{britannicaAlgaePhotosynthesis}. 
This description of algae is under the wider scope of alga definition. 
Cyanobacteria (blue-green algae) are considered as microscopic algae and are widely distributed in aquatic environments across many climate zones. 
Cyanobacteria blooms may produce cyanotoxins, generate taste-and-odor problems, use large amounts of oxygen in the water and/or create surface scums that degrade water quality. These blooms can be detrimental to public health and are referred to as \acfp{HAB}. 

\begin{figure}
    \centering
        \includegraphics[width=0.75\linewidth]{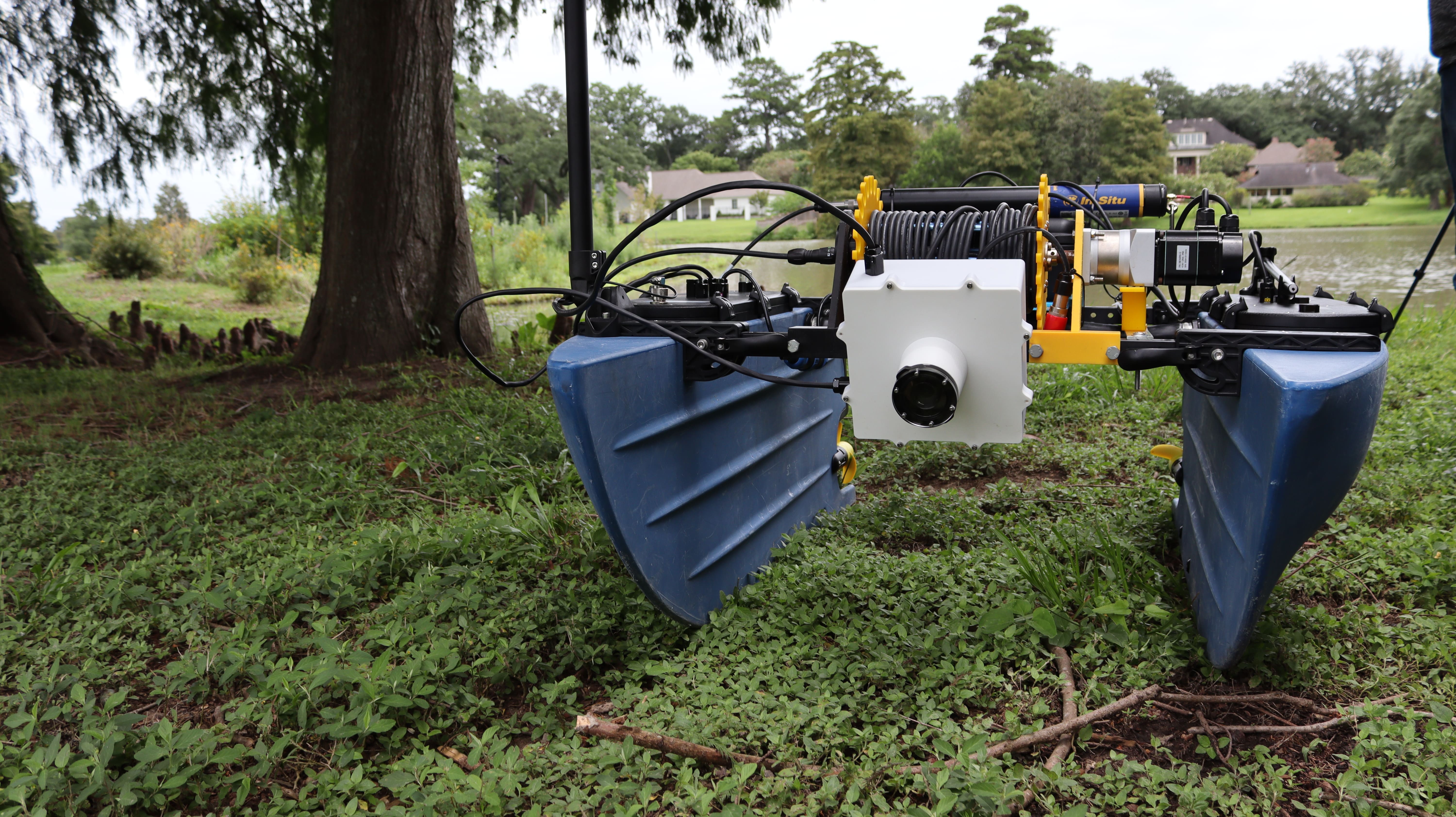}
    \captionsetup{hypcap=false}
    \captionof{figure}{\acf{ASV} designed for \acf{HAB} monitoring}
    \label{fig:mainsystem}
    \vspace{-15pt}
\end{figure}

Because bloom formation is highly dynamic spatially and temporally, continuous monitoring is essential to support early detection, risk assessment, and timely response. Given the large spatial extent of water bodies, conventional point-based sampling may require an impractically dense sampling network to capture fine-scale variability. This can force a trade-off toward lower spatial resolution. Autonomous robotic monitoring can provide a practical alternative by enabling repeatable high-resolution measurements over such areas, improving the ability to detect, track, or predict bloom formation. Such platforms, including \acp{UAV} and \acp{ASV}, have been leveraged for environmental monitoring and marine data collection in recent years \cite{YUAN2023159741}, but have not been explored for \ac{HAB} monitoring. To address this limitation, this paper presents an autonomous surface vehicle system for algal bloom monitoring and \ac{HAB} prediction that integrates repeatable in situ sensing with machine-learning-based identification of \ac{HAB} formation  (\figurename~\ref{fig:mainsystem}). By combining robotic data acquisition with data-driven analysis, the proposed approach aims to improve spatially resolved monitoring and support early assessment of bloom formation in natural water bodies.

\vspace{-5pt}
\section{Related Work}
\label{sec:related-work}
A review of the application of \acp{ASV} in marine environmental monitoring \cite{LIU2025102926} noted the sharp increase in environmental monitoring studies in the last 10 to 15 years in response to water pollutants, climate change, and other dynamic environmental factors. Across these studies, \acp{UAV} dominated the research field, being the primary tool of data collection in 70\%. \acp{UAV} are well suited for imaging at medium to large spatial scales, even in aquatic environments that \acp{ASV} cannot reach due to shallow waters or surface clutter. Some can even land on the water's surface and take precise point-to-point measurements. However, \acp{UAV} suffer from much shorter mission times due to reduced battery capacity compared to \acp{ASV} and are limited to surface and shallow water measurements \cite{w10030264}. \acp{ASV} are better suited for applications requiring extended mission times and subsurface measurements. Still, they represent an under-utilized tool in the field of aquatic research.

In studies utilizing \acp{ASV}, algal bloom monitoring is frequently a desired capability. The works presented in \cite{Dunbabin2009ASVWaterQuality}, \cite{s25123747}, and \cite{Yaakob2012USVSeaPatrol} have developed \acp{ASV} suitable for these missions, each designed for deployment in different environments. These designs are large and costly, so seeing the need for miniaturization, \cite{JO2019e00076} developed a smaller \ac{ASV}. However, most of these platforms, with the exception of \cite{Dunbabin2009ASVWaterQuality}, have rigid sensors and are limited to shallow water. This leaves a need for compact, affordable \acp{ASV} with sensor packages that are deployable to selectable depths, enabling a richer understanding of algal bloom behavior throughout the water column.

A common application of the water quality data gathered by these vehicles is the modeling and prediction of \acp{HAB}, often using data-driven machine learning approaches. The AlgAlert program \cite{ALSINI2025103260}  was created to evaluate the performance of six neural network training algorithms when predicting Chlorophyll-a concentrations, and then subsequently classifying \acp{HAB}. 

A Pearson correlation matrix was used in \cite{ML-Extrusion} to reduce the number of input features and increase accuracy of produced machine learning models.

\vspace{-3pt}
\section{Methodology}
\label{sec:methodology}
In this paper, the methodology is organized around two main components: (1) autonomous in-situ environmental data collection using an \ac{ASV}-based sensing platform and (2) identification of algae blooms based on data collection and machine learning methods. The primary system components of the sensing platform include the Blue Robotics BlueBoat \cite{blueroboticsBlueBoatUncrewed}, a motorized winch system, a multiparameter water-quality sonde, a side-scan sonar system, and an event-camera system. Together, these components enable autonomous navigation to predefined sampling locations, collection of water-quality measurements at multiple depths, and recording of complementary environmental data during each mission.
The machine learning model that classifies \ac{HAB} consists of two layers. The first layer predicts the concentration of chlorophyll-a in a given area. The second layer uses a predetermined chlorophyll-a concentration threshold to classify the measurements as \ac{HAB}. The following paragraphs describe in detail the proposed approaches. 

\subsection{Platform, Instruments, and Sensors}
\vspace{-3pt}
\subsubsection{Autonomous Surface Vessel (ASV)}
The proposed system was designed to enable affordable autonomous in-situ collection of water quality data relevant to algal bloom formation in natural water bodies, with emphasis on repeatability, modularity, and scalability. To support autonomous deployment of the sensing payload, all instruments and sensors were mounted on a BlueBoat \ac{ASV} from Blue Robotics \cite{blueroboticsBlueBoatUncrewed}, shown in \figurename~\ref{fig:mainsystem}.  The platform provided a practical base, since the sensing payloads could be integrated without substantially modifying the baseline propulsion, communication, and navigation systems. The BlueBoat features a modular payload architecture with multiple available mounting points, internal hull space for electronics, dedicated payload power, integrated \ac{GPS}, and open-source onboard software (BlueOS) \cite{blueroboticsBlueOS} with autonomous \ac{GPS} waypoint navigation available via ArduRover \cite{ardupilotArduPilotRover}, allowing execution of predefined survey missions. 

\begin{figure}
    \centering
    \begin{subfigure}[t]{0.4\linewidth}
        \centering
        \includegraphics[width=\linewidth]{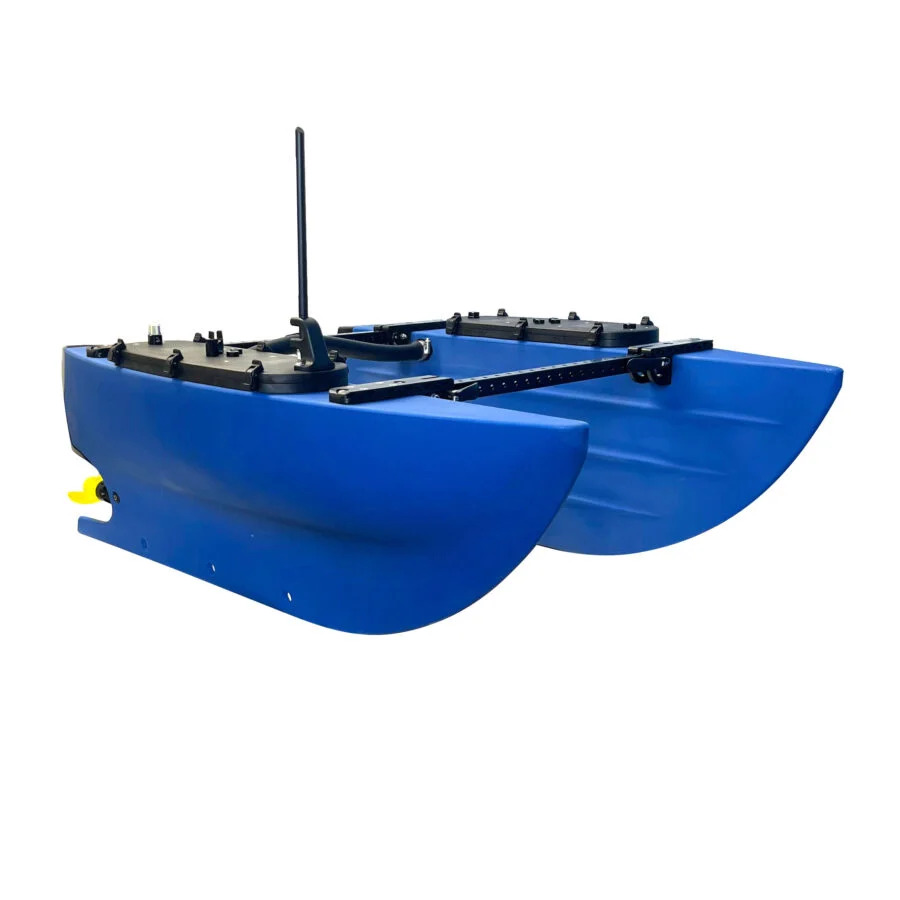}
        \caption{}
    \end{subfigure}
    \hfill
    \begin{subfigure}[t]{0.56\linewidth}
        \centering
        \includegraphics[width=\linewidth]{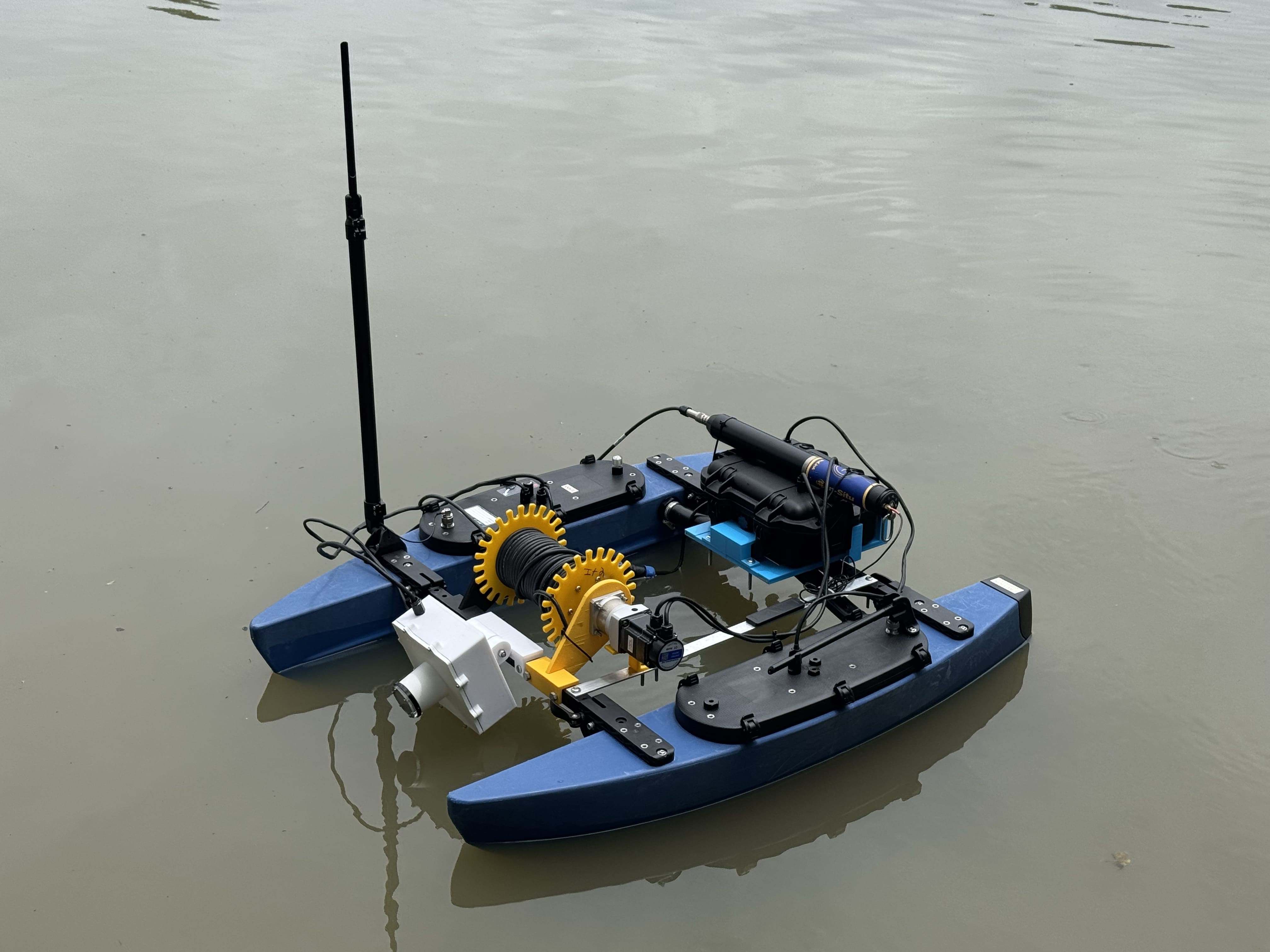}
        \caption{}
    \end{subfigure}
    \caption{(a) Baseline BlueBoat \ac{ASV} platform prior to payload integration (\textit{Image credit: Blue Robotics})  and (b) after installation of the integrated sensing and actuation subsystems}
    \label{fig:beforeandafter}
\end{figure}

These features make it well suited for integrating environmental sensors and performing repeatable autonomous data-collection missions. The payload architecture serves as the anchor point for additional sensing and actuation subsystems: the winch, water-quality sensing, event camera, and sonar payloads used in this work. The use of autonomous waypoint navigation further enables repeatable surveys, allowing the \ac{ASV} to revisit the same sampling sites to execute the same paths across multiple deployments. This repeatability is important for monitoring spatial and temporal changes in the water-quality conditions associated with algal bloom development. \figurename~\ref{fig:beforeandafter} highlights the state of the \ac{USV} before and after installation of the integrated subsystems.

\vspace{2pt}
\subsubsection{Winch System}
A winch system was designed and integrated with the BlueBoat to enable in-situ water-quality measurements at specific depths. The winch lowers and raises the sonde during deployment, allowing measurements to be collected at different depths without manual intervention. Actuation of the winch is provided by a NEMA 23 stepper motor, selected for its ability to provide sufficient holding torque for the suspended sonde. The motor is driven by a stepper driver connected to the vehicle's power system, with the sonde depth determined by a calibrated step-to-depth conversion.

To meet the torque and speed requirements of the winch system, additional mechanical and power modifications were required. The aforementioned modifications include a gearbox and a boost converter. Although the stepper motor could operate on a 15V supply, this voltage did not provide sufficient torque for reliable lifting and lowering of the sonde. The boost converter was therefore added to increase the voltage provided to the stepper driver, resulting in around 35V in total to the motor. The stepper gained a sufficient amount of power to hold, lift and lower the sonde; however, the speed was not sufficient considering mission time constraints. This was due to the booster converting current into voltage. As a result, a 5:1 gearbox was added to the stepper motor. This increased the effective output torque, allowing the supply voltage to be reduced while maintaining sufficient torque to lift, lower and hold the sonde during autonomous sampling.

\begin{figure}
    \centering
    \begin{subfigure}[t]{0.45\linewidth}
        \centering
        \includegraphics[width=\linewidth]{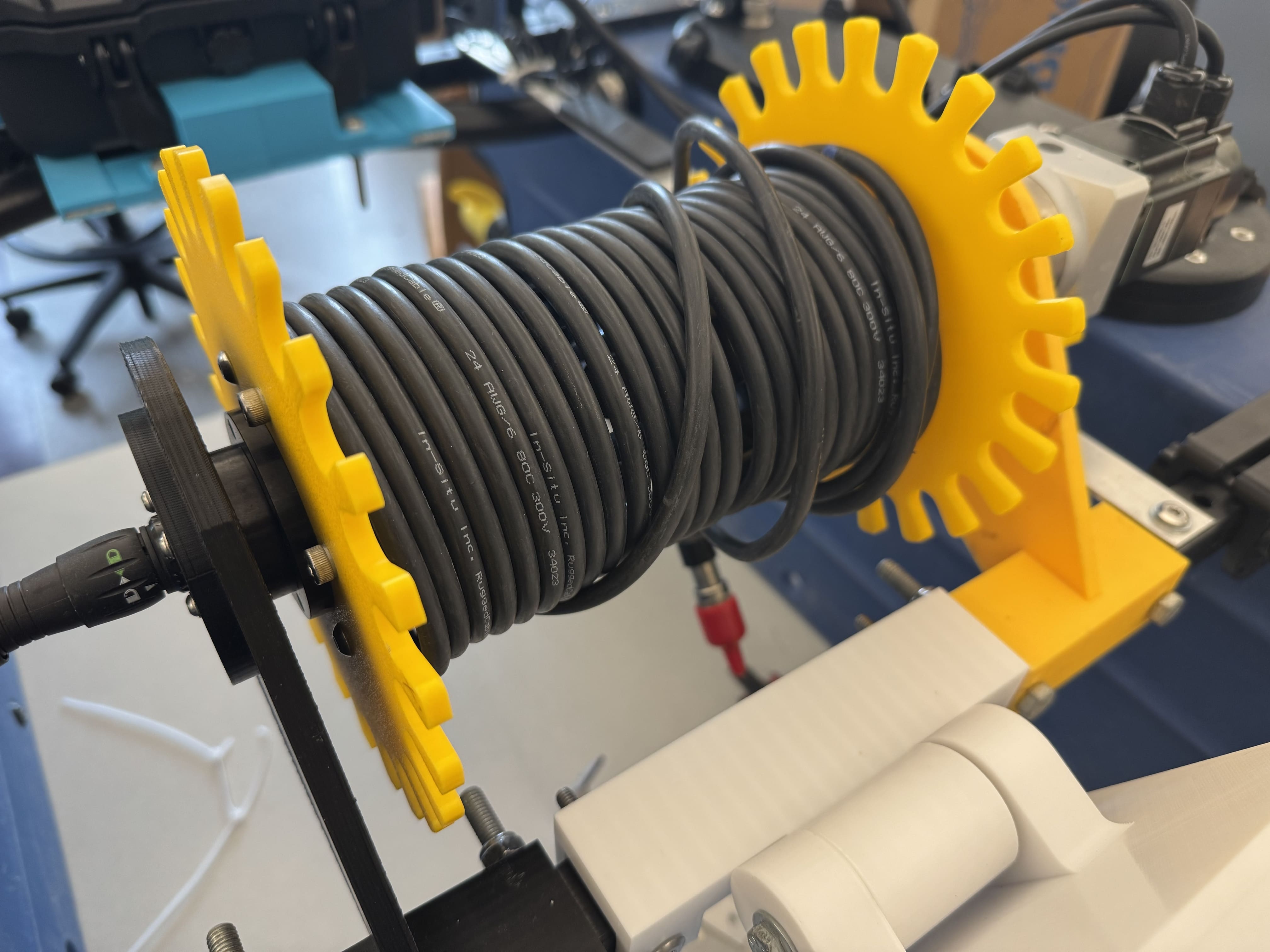}
        \caption{}
        \label{fig:winchphys}
    \end{subfigure}
    \begin{subfigure}[t]{0.45\linewidth}
        \centering
        \includegraphics[width=\linewidth]{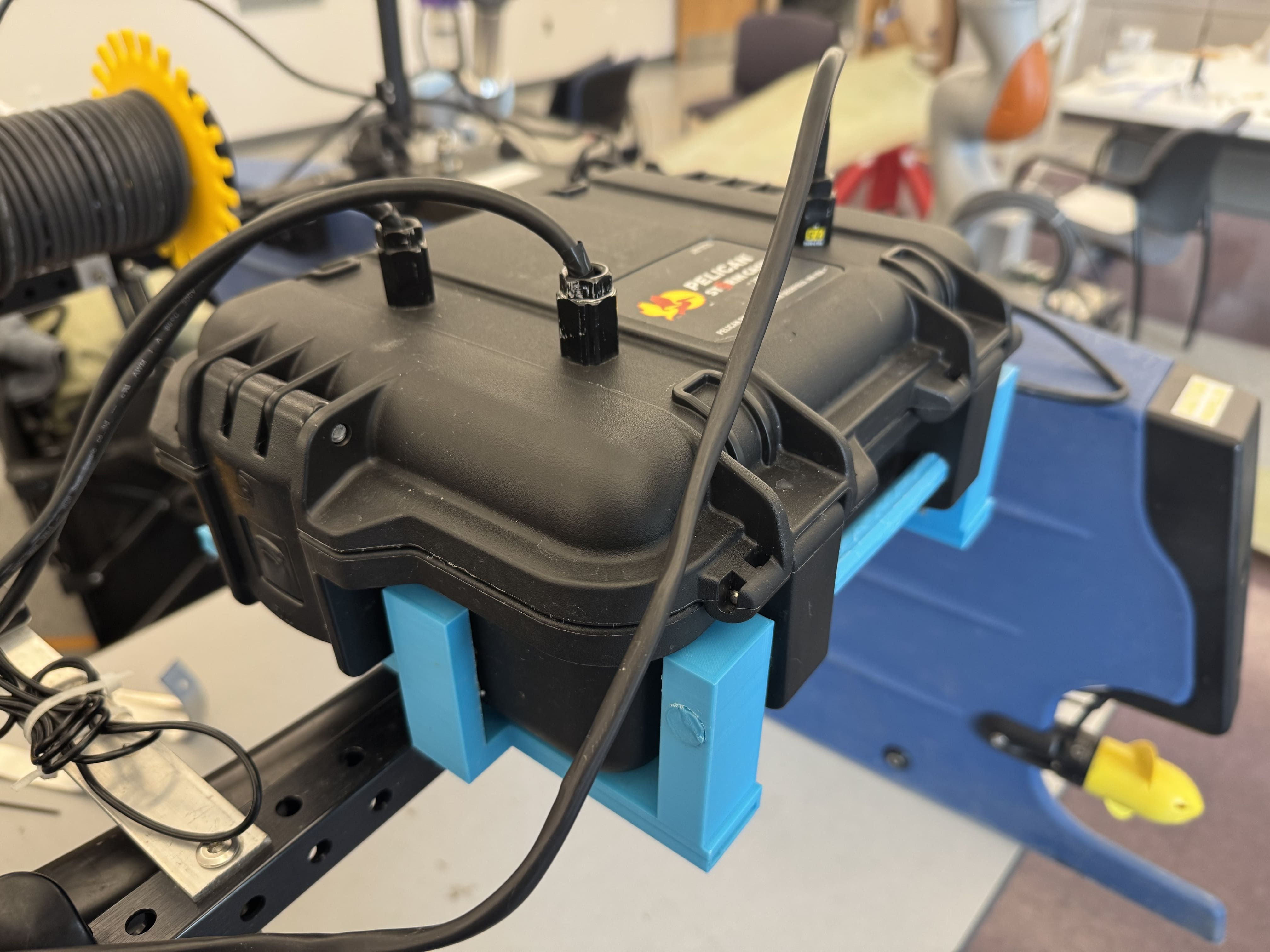}
        \caption{}
        \label{fig:winchcase}
    \end{subfigure}
    \caption{Integrated winch system installed on the \ac{ASV}, showing the (a) mechanical assembly and (b) associated electronics enclosure}
    \label{fig:winchsystem}
\end{figure}

The winch assembly is attached to the payload mounting frame of the \ac{ASV} and positioned to allow the sonde cable to be deployed vertically in the water. The sonde cable is wound around a spool attached to the motor output shaft as shown in \figurename~\ref{fig:winchphys}.

The winch is controlled through the on-board computing system, which sends depth commands to an Arduino microcontroller, housed in a separate casing, \figurename~\ref{fig:winchcase},  responsible for low-level motor actuation. The microcontroller converts the requested depth to the corresponding number of motor steps and generates the required step and direction signals for the stepper driver. This task separation allows the on-board computer to handle mission-level commands only while the microcontroller performs timing-sensitive motor control. 

\vspace{2pt}
\subsubsection{Multiparameter Sonde}
\begin{figure}
    \centering
    \begin{subfigure}[t]{0.48\linewidth}
        \centering
        \includegraphics[width=\linewidth]{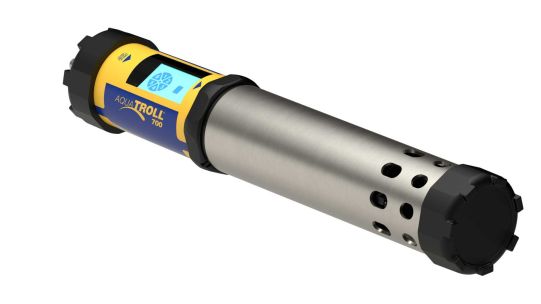}
        \caption{\textit{Image credit: In-Situ}}
        \label{fig:aquatroll}
    \end{subfigure}
    \hfill
    \begin{subfigure}[t]{0.48\linewidth}
        \centering
        \includegraphics[width=\linewidth]{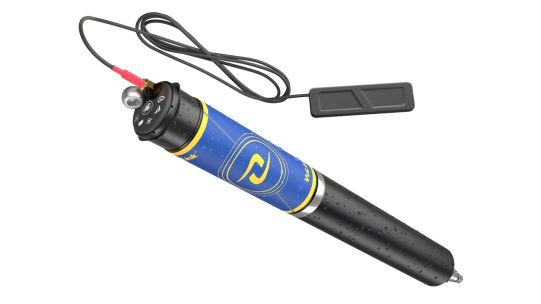}
        \caption{\textit{Image credit: In-Situ}}
        \label{fig:vulink}
    \end{subfigure}
    \caption{Water-quality sensing components: (a) Aqua TROLL 700 multiparameter sonde and (b) VuLink cellular telemetry device}
    \label{fig:sondesystem}
\end{figure}

The primary sensing payload for water-quality measurements relevant to algal bloom formation was the Aqua TROLL 700 multiparameter sonde (\figurename~\ref{fig:aquatroll} \cite{InSitu2024AquaTROLL}). The sonde was configured with six individual sensors from which key water-quality parameters are directly obtained or derived: Nitrate, Chlorophyll-a,  \acf{BGA-PC}, Temperature,  Turbidity, \ac{FDOM}. 

These sensors were selected to capture key biological and environmental indicators associated with algal bloom development. Nitrate indicates nutrient availability, chlorophyll-a fluorescence provides a proxy for algal biomass, and BGA-PC fluorescence provides an indicator of cyanobacteria-related activity. Temperature, turbidity, and \ac{FDOM} provide additional context regarding thermal conditions, water clarity, suspended material, and dissolved organic matter, all of which influence water-quality conditions associated with bloom formation \cite{Johnston2024}. 

The sonde is connected through a slip ring to a dedicated communication module, the VuLink Cellular Telemetry Device (\figurename~\ref{fig:vulink}), which provides power to the sonde and transmits the collected measurements to a cloud-based storage. This configuration allows the sonde to remain powered while having access to live measurements on-shore during missions.

\vspace{2pt}
\subsubsection{Side-scan Sonars}
Following integration of the multiparameter sonde, a sonar system was installed on the \ac{ASV}. The system consists of two Cerulean Omniscan 450 \ac{SS} sonars mounted externally on the port and starboard hulls of the \ac{ASV}. The sonar transmits acoustic pulses in narrow, vertically oriented, fan-shaped beam patterns, and the reflections of these pulses provide information on the texture and reflectivity of the surrounding terrain, as well as any sufficiently large features or objects present in the environment. As the vehicle navigates, the system uses GPS-based and onboard IMU-based position and heading to map the data into a 2D image. The result is an acoustic image of the marine environment around the \ac{ASV} that is immune to attenuation and degradation in turbid water. 

A sample \ac{SS} acoustic image gathered from tests at False River is shown in \figurename~\ref{fig:sidescan}. The granular image quality is characteristic of backscattered sonar pulses, and without sufficiently large ground clutter, most of the image is featureless. However, several long, thin features (possible rope at the bottom of the lake) and a change in lake bed texture are clearly visible in the image, indicating the system's ability to gather useful information despite a poor \ac{SNR}.
\vspace{2pt}
\subsubsection{Event Camera}
\begin{figure}
    \centering
    \begin{subfigure}[t]{0.4\linewidth}
        \centering
        \includegraphics[width=\linewidth]{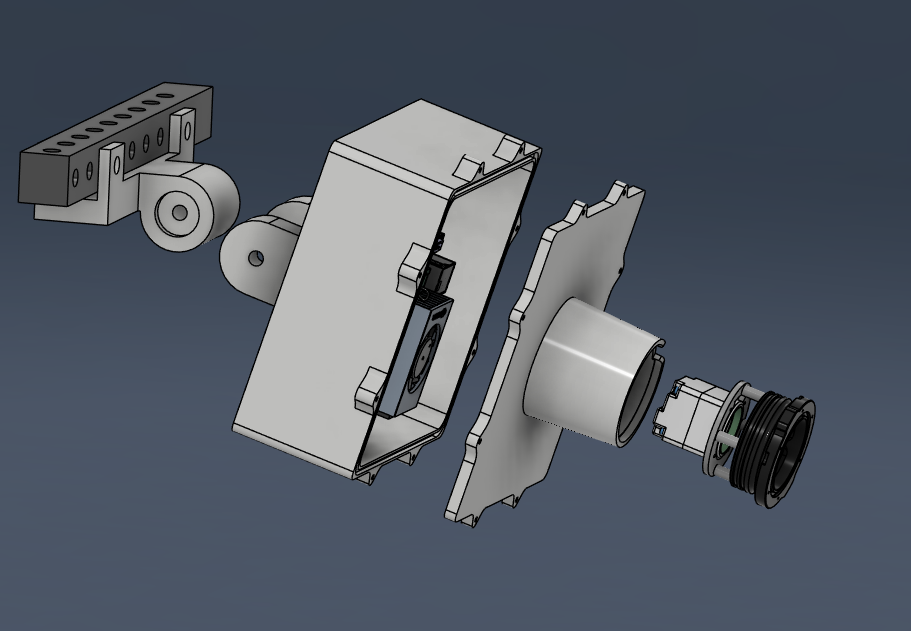}
        \caption{}
        \label{fig:housingexploded}
    \end{subfigure}
    \begin{subfigure}[t]{0.48\linewidth}
        \centering
        \includegraphics[width=\linewidth]{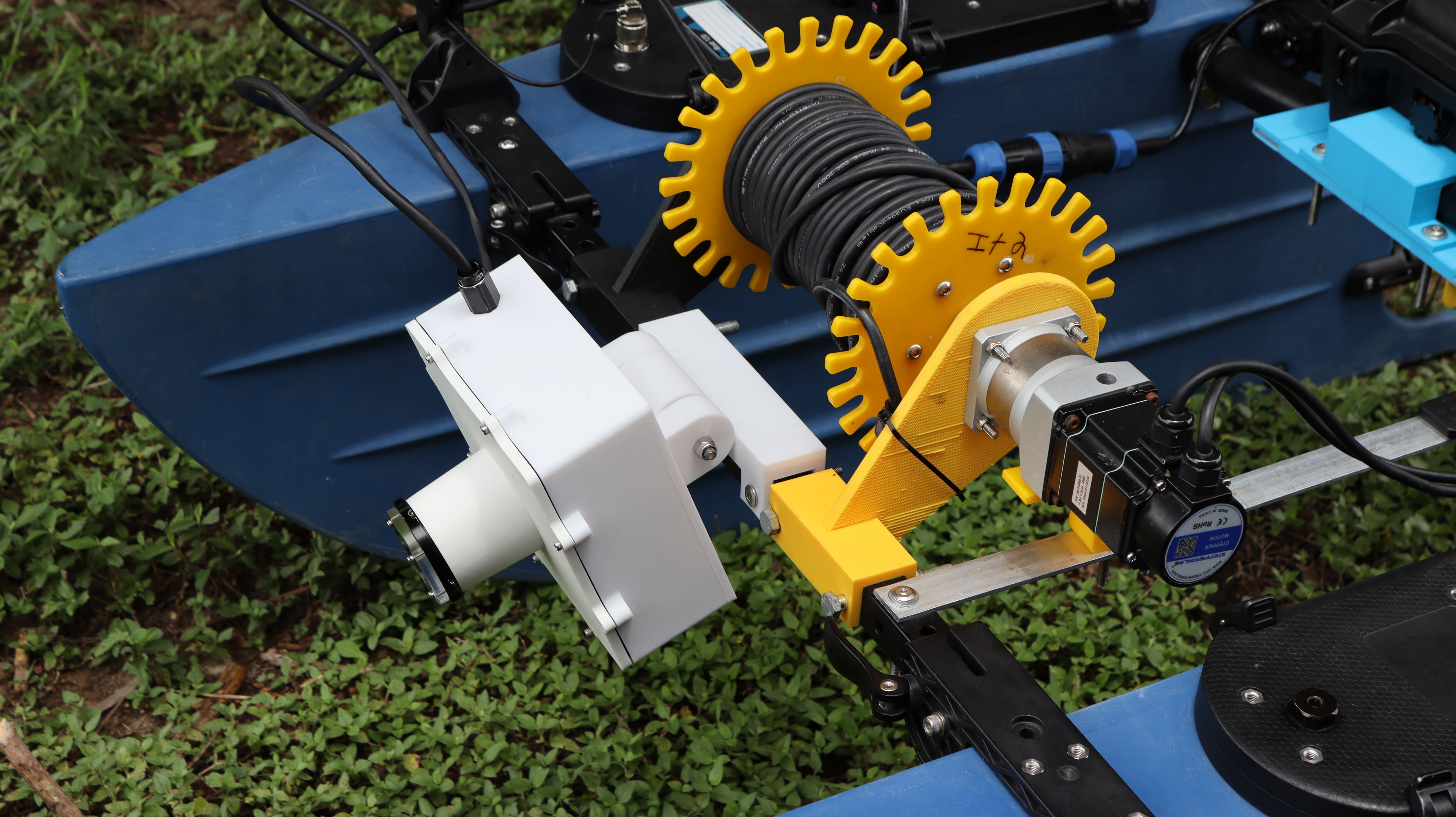}
        \caption{}
        \label{fig:cameracasereal}
    \end{subfigure}
    \caption{(a) CAD model of event camera system and its (b) physical mounting.}
    \label{fig:eventcamerasystem}
\end{figure}

The final system implemented on the \ac{ASV} was an event camera system. The system was housed in an external enclosure containing the camera and an NVIDIA Jetson Orin Nano Super Developer Kit running Ubuntu \ac{OS}. The housing utilized off-the-shelf components from BlueRobotics and custom 3D-printed Polyethylene Terephthalate Glycol (PETG) components. A spliced O-ring sealed the interface between body and lid while additional O-rings on the BlueRobotics flange helped seal the bore where the camera was mounted. BlueRobotics cable penetrators were used to seal the cable interface. Prototype submersion testing confirmed that the housing was largely water-resistant, though prolonged submersion resulted in slow leaks due to water-permeable 3D-printed parts and poor sealing between the O-rings and their contact surfaces. This was deemed suitable for deployment on the \ac{ASV}, where the event camera would not be exposed to deep underwater environments. \figurename~\ref{fig:housingexploded} shows the exploded CAD assembly of this system, without the cable penetrators, which were machined after-the-fact while \figurename~\ref{fig:cameracasereal} shows the deployed 3D-printed prototype version of the enclosure.

The Jetson was connected to the \ac{ASV} through cables routed from the housing into the \ac{ASV}'s starboard hull. These cables included a power cable and an Ethernet cable, which enable communication between the \ac{ASV} Raspberry Pi and the Jetson through \ac{SSH}. This connection allowed the Jetson to be accessed from the \ac{ASV} onboard network for event camera operation and data management.

\vspace{-7pt}
\subsection{General Electrical System}
The \acf{ASV} integrates multiple additional subsystems alongside the base platform components responsible for propulsion, communication, navigation, and localization. These added subsystems include the winch system, multiparameter sonde, sonar system, and event camera system. All systems, except the multiparameter sonde, are connected through the \ac{ASV}'s in-hull primary processing unit (Raspberry Pi), through various communication peripherals such as standard serial, I2C, and Ethernet connections. 

The added subsystems, barring the multiparameter sonde, share the vehicle power-distribution architecture. Power is supplied by Blue Robotics Lithium-ion batteries distributed between the two hulls to maintain vehicle balance. Each battery has a 4S6P cell configuration, providing a nominal voltage of 14.8V with a nominal capacity of 15.6Ah. Electrical power is distributed between both \ac{ASV} hulls through power-splitting cables and routed through the fuse boards housed within each hull. Each subsystem remains inactive until the appropriate fuse is installed on the fuse board, providing a layer of protection during installation, testing, and field operation.

\vspace{-5pt}
\subsection{Autonomous Data Collection and Mission Structure}
The autonomous measurement strategy is based on a sequence of predefined waypoints distributed across the study area. These waypoints define the locations where in-situ measurements are collected, while the specific arrangement of the mission can be adjusted according to the survey objective, site geometry, and desired spatial coverage. The \ac{ASV} autonomously navigates between the selected waypoints using the \ac{GPS} navigation, enabling repeatable missions without manual vehicle control. 

The system collects measurements at multiple depths using the motorized winch system described previously. Upon reaching a waypoint, the on-board computer sends a command over serial to the winch to lower the multiparameter sonde to specified depths. The Arduino microcontroller converts the commanded depth into the corresponding number of motor steps using the calibrated step-to-depth relationship and generates the stepper motor control signals required for deployment and retrieval. This allows multiple subsurface measurements to be collected during a single deployment without manual intervention.

At each sampling point, the following environmental parameters are measured to characterize conditions relevant to algal bloom development: Chlorophyll-a Fluorescence, FDOM Fluorescence, Turbidity, Conductivity, Salinity, Resistivity, Density, Total Dissolved Solids, BGA-PC Fluorescence, Nitrate concentration, Temperature, Barometric Pressure, Pressure, Depth, and Voltage. These measurements are associated with the corresponding waypoint, sampling depth, and collection time so that water-quality conditions can be compared across sampling locations and repeated missions. Measurements are recorded and stored simultaneously on the VuLink Cellular Telemetry Device and cloud-based storage provided by the the VuLink service.

In addition to the discrete sonde measurements, the sonar and event camera systems are initialized at the start of the mission to collect continuous data during vehicle operation. Event camera recordings are stored directly on the NVIDIA Jetson, while sonar data is collected, both on the Raspberry Pi and the ground control station, to provide complementary information about the surrounding environment. Together, the waypoint mission, depth-controlled sonde deployment, and continuous sensor recordings provide a repeatable framework for collecting spatially distributed, multi-modal environmental data relevant to algal bloom monitoring.
\vspace{-7pt}
\subsection{Algal Bloom Classification}
The machine-learning approach developed for \ac{HAB} classification is comprised of two components: a Random Forest prediction layer and a binary classification layer. In the first stage, the Random Forest model predicts chlorophyll-a concentration using a combination of environmental and sensor derived variables including: external voltage, elevation, longitude, latitude, depth, nitrate concentration, temperature, turbidity, barometric pressure, specific conductivity, BGA-PC fluorescence, and FDOM fluorescence. The input variables are chosen by passing all possible features through a Pearson correlation matrix, and removing the features with high correlations to others as done in \cite{ML-Extrusion}. After chlorophyll-a concentration is predicted, the classification layer uses a predetermined threshold to determine if the measurement is a \ac{HAB}. Any predicted chlorophyll-a value greater than 10 $\mu$g/L is labeled as a \ac{HAB} as is consistent with the approach used in \cite{ALSINI2025103260}.

\vspace{-10pt}
\section{Experiment \& Results}
\label{sec:results}
This section presents the field deployments used to evaluate the autonomous sensing system and methods, including the mission setup, collected water-quality measurements, complementary sensor data and machine learning model performance. 
\vspace{-3pt}
\subsection{Field Deployment and Mission Setup}
Primary data collection was conducted in Lake Pontchartrain, Louisiana. Lake Pontchartrain is one of the major estuaries of the east Mississippi Delta. The combination of warm temperatures and influx of the Mississippi River nutrients reaching the lake from the Bonnet Carre Spillway contributes to \ac{HAB} formation in the lake \cite{nasaColoringLake}.

\begin{figure}
    \begin{subfigure}[t]{0.49\linewidth}
    \centering
    \includegraphics[width=0.9\linewidth]{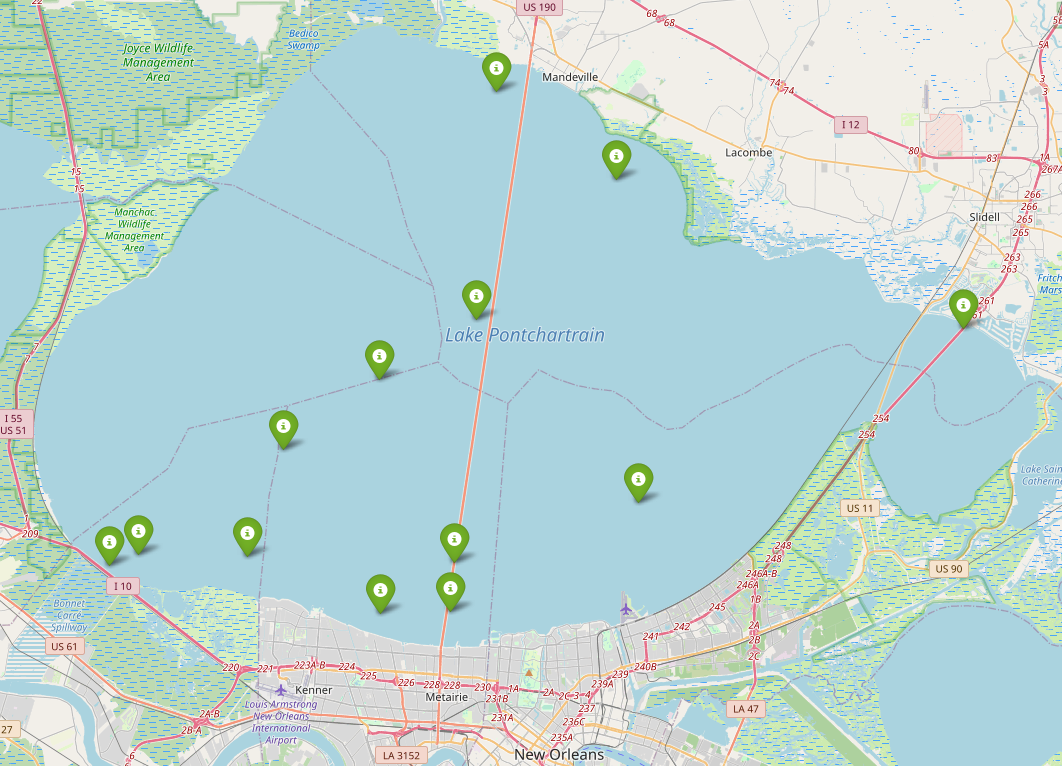}
    \caption{Map of sampling sites on Lake Pontchartrain}
    \label{fig:mappoints}
    \end{subfigure}
 \begin{subfigure}[t]{0.45\linewidth}
    \centering
    \includegraphics[width=0.9\linewidth]{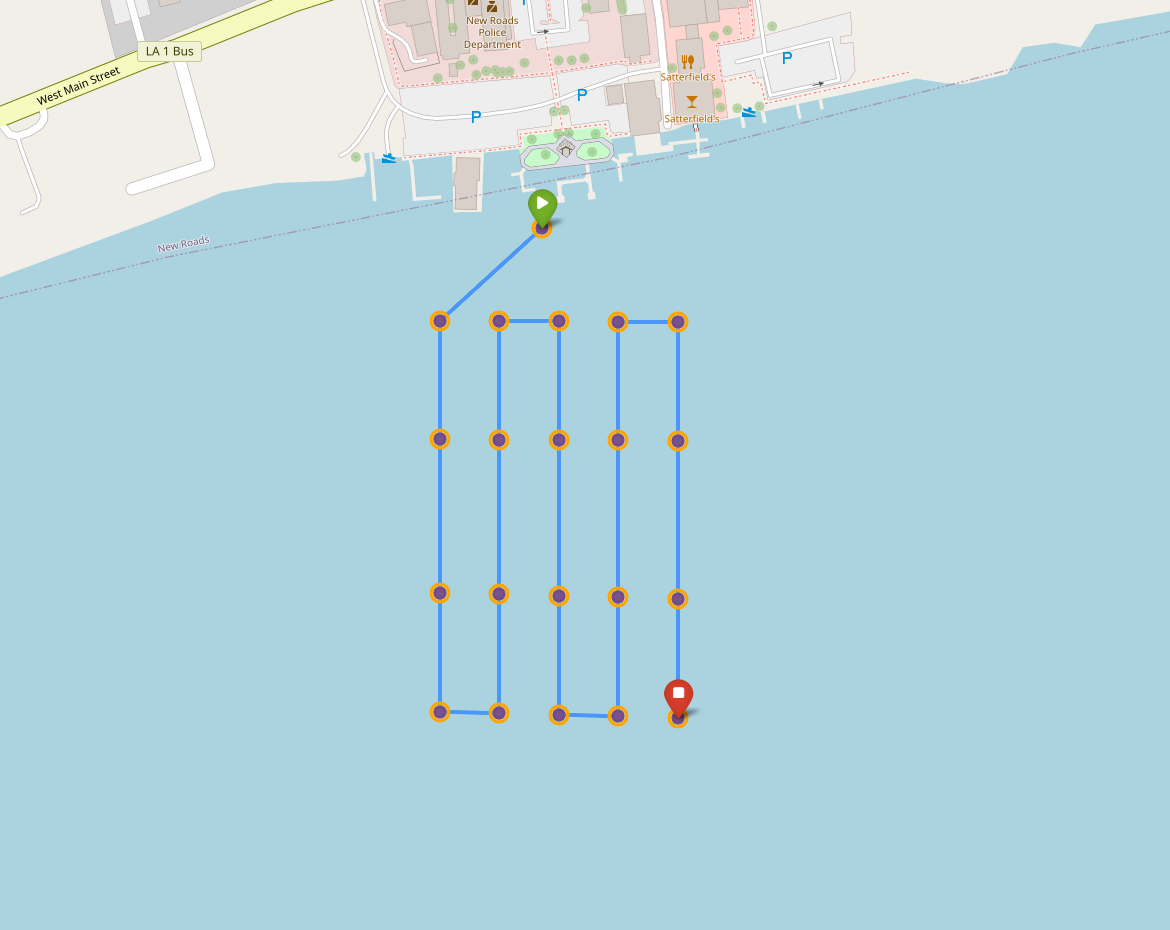}
    \caption{Lawnmower path and waypoints used on False River}
    \label{fig:falseriverpath}
    \end{subfigure}
\end{figure}

In \figurename~\ref{fig:mappoints} each pin denotes a distinct sampling site. The autonomous system was deployed at select areas off the northern and southern coasts of the lake. 

At each site, the vehicle performed a localized lawnmower coverage pattern to obtain multiple samples within the region, providing improved spatial resolution compared with a single point measurement. In addition to horizontal coverage, measurements were autonomously collected at multiple depths ($0.5$m, $1$m, and $2$m) through the motorized-winch mechanism that lowers the environmental sensor. This combined sampling strategy enables both horizontal and vertical characterization of environmental parameters associated with algal bloom development.

During mission execution, sonde deployment was coordinated with the autonomous waypoint mission using an onboard Lua script. The script monitored the currently active waypoint in the uploaded mission and triggered the winch system when the \ac{ASV} reached a waypoint designated for sampling. At each sampling point, the vehicle was placed in loiter mode while the Arduino-controlled winch performed the data collection sequence. After the Arduino reported that data collection was complete, the waypoint was marked as collected and the vehicle returned to autonomous waypoint navigation. This allowed the \ac{ASV} to pause at each sampling location, collect depth-controlled measurements, and continue the mission without manual intervention.

Although Lake Pontchartrain served as the primary study site, additional data were collected in False River to expand the spatial diversity of the dataset. These supplementary measurements provided observations from a separate aquatic environment, supporting broader modeling and investigation of algal bloom conditions. Unlike the strategy adopted for Lake Pontchartrain, a single data collection site was chosen 
as shown in \figurename~\ref{fig:falseriverpath}.

\begin{figure}
    \centering
    \includegraphics[width=0.8\linewidth]{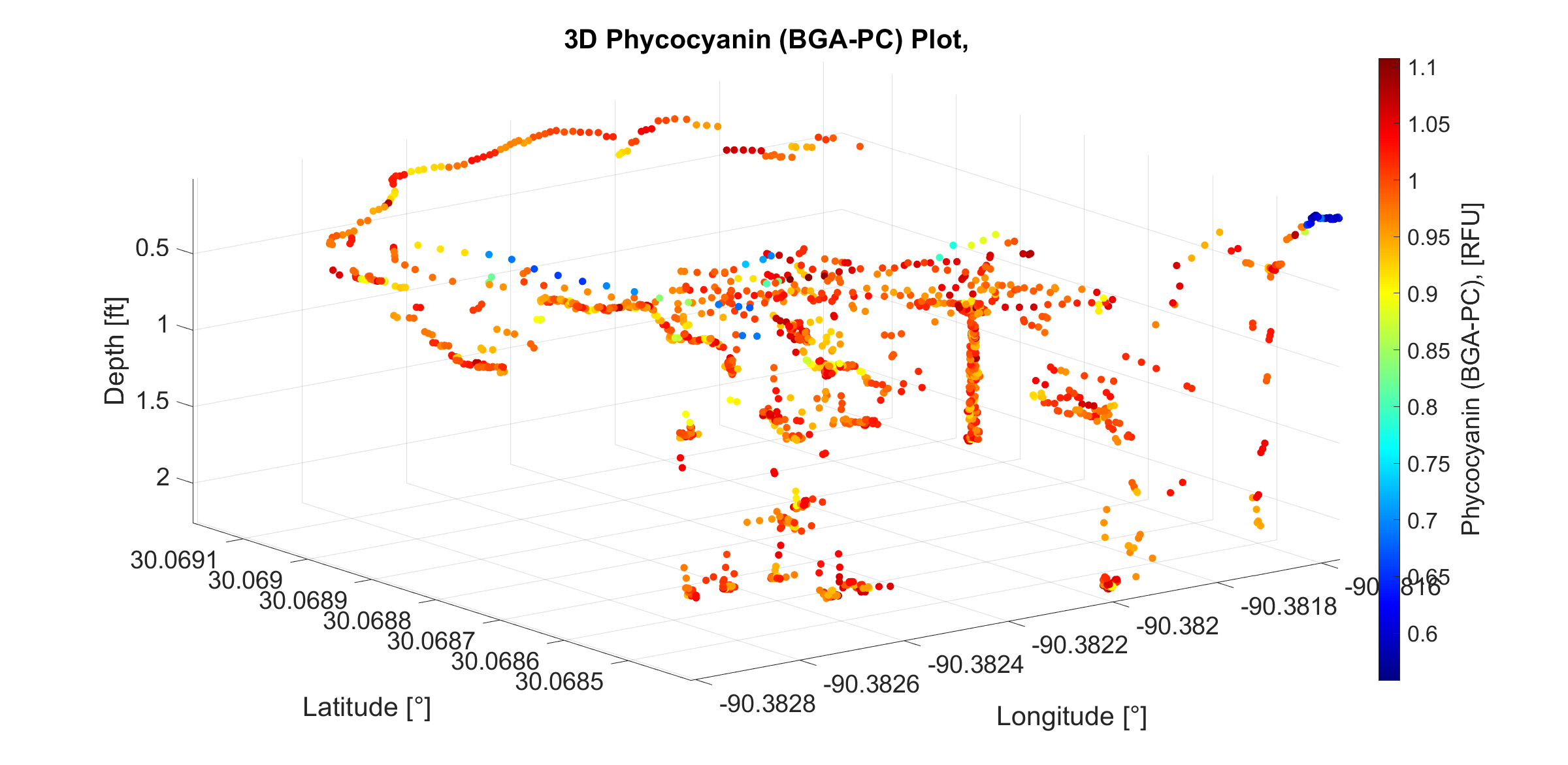}
    \caption{Data collected with Jenny \ac{ASV}}
    \label{fig:data_collected}
    \vspace{-5pt}
\end{figure}

\begin{figure}
    \centering
    \captionsetup{justification=centering}
    \includegraphics[width=0.9\linewidth]{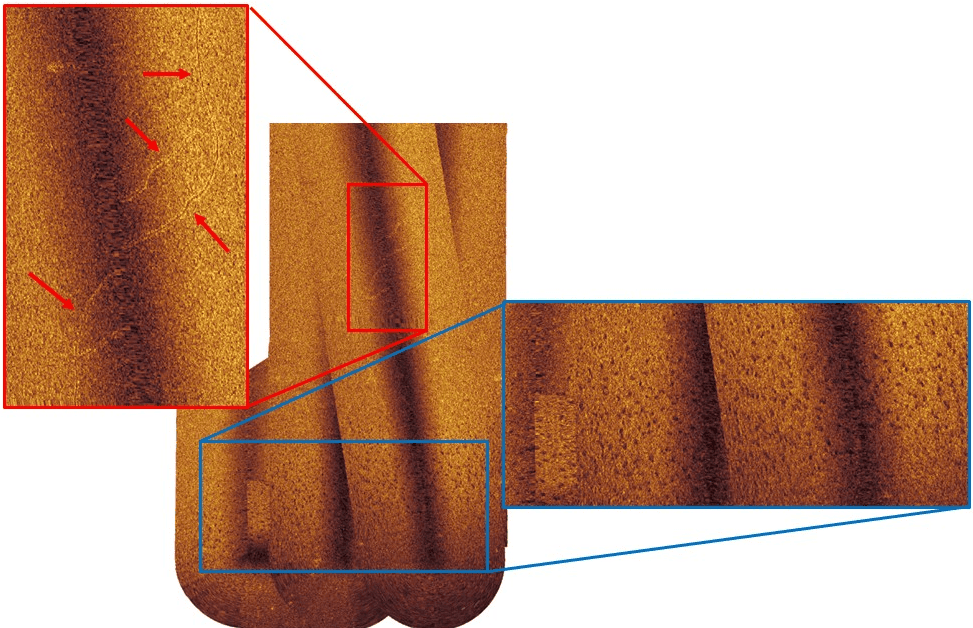}
    \caption{Acoustic image from side scan sonar, highlighting feature and texture imaging}
    \label{fig:sidescan}
\end{figure}

\begin{figure*}[!t]
    \centering
    \begin{subfigure}[t]{0.24\textwidth}
        \centering
        \includegraphics[width=\linewidth]{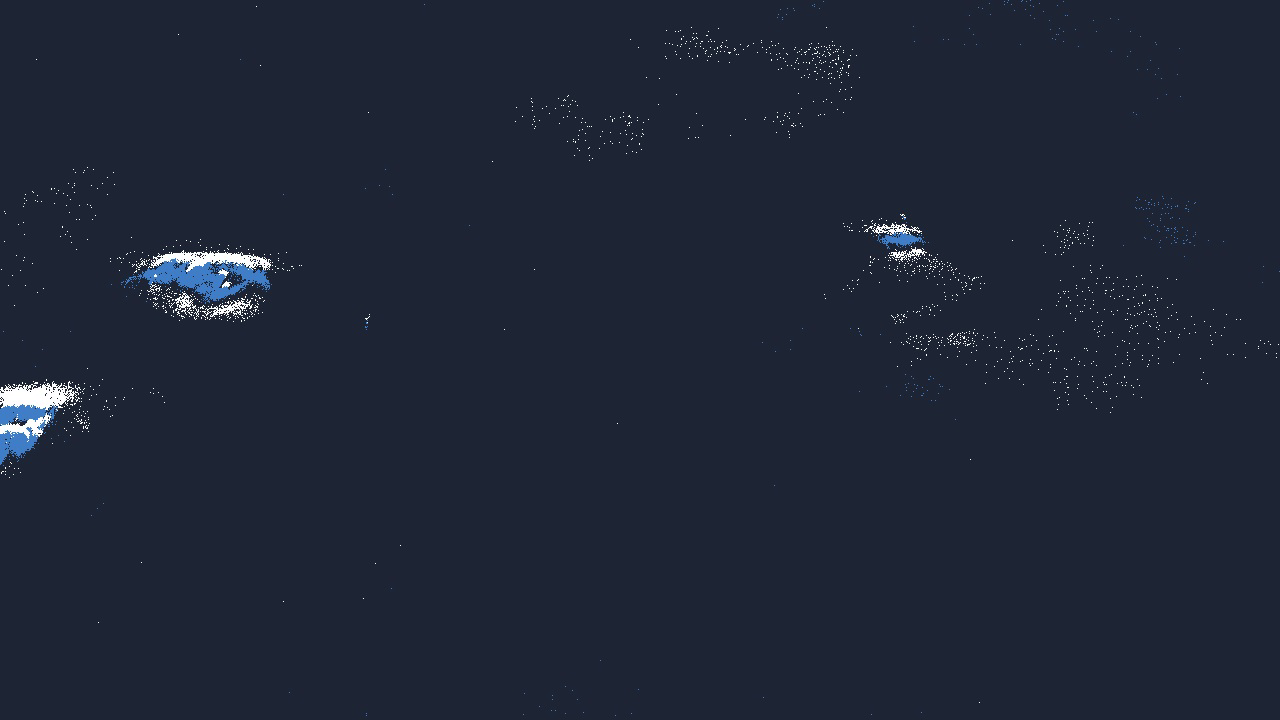}
        \label{fig}
    \end{subfigure}
    \hfill
    \begin{subfigure}[t]{0.24\textwidth}
        \centering
        \includegraphics[width=\linewidth]{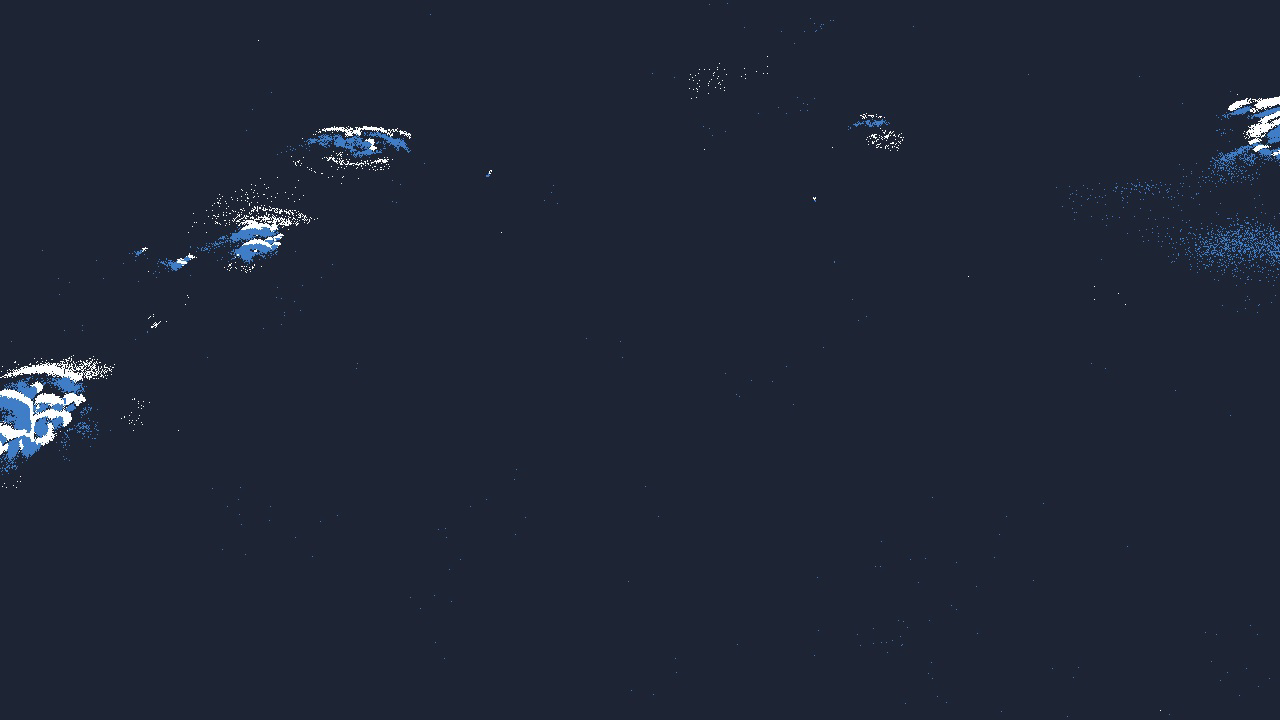}
        \label{fig}
    \end{subfigure}
    \hfill
    \begin{subfigure}[t]{0.24\textwidth}
    \centering
    \includegraphics[width=\linewidth]{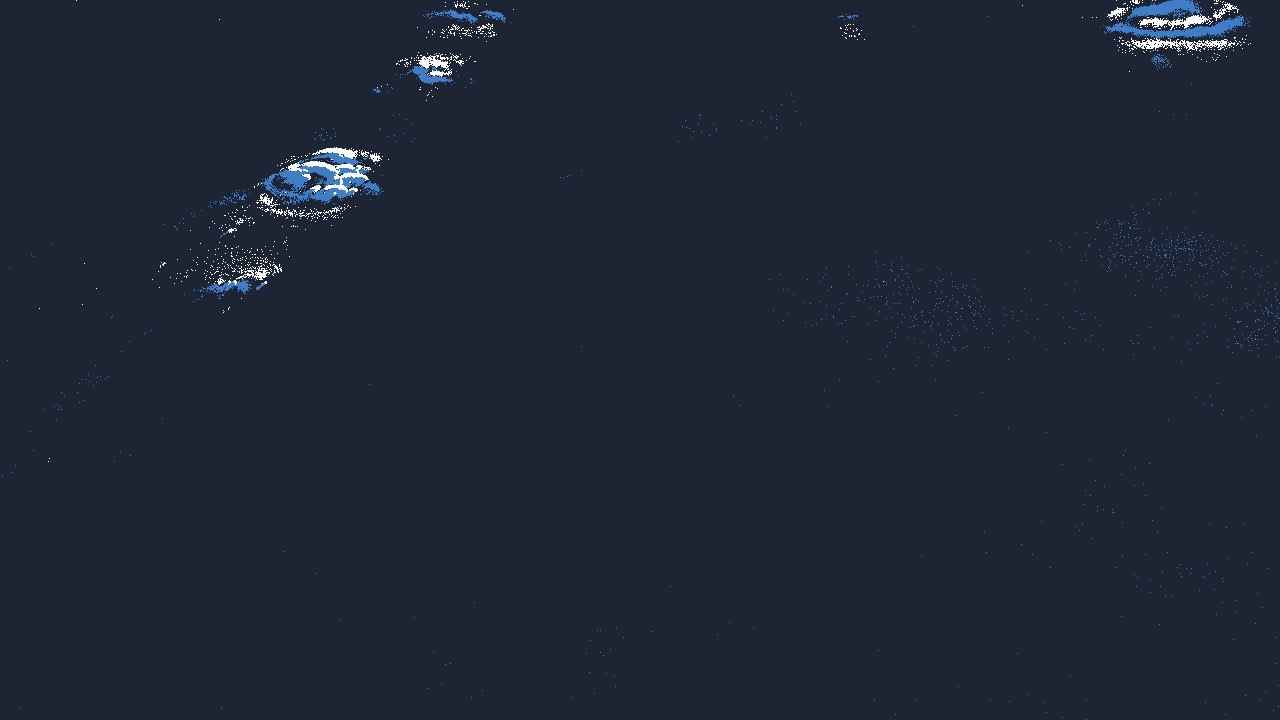}
    \label{fig}
    \end{subfigure}
    \hfill
    \begin{subfigure}[t]{0.24\textwidth}
    \centering
    \includegraphics[width=\linewidth]{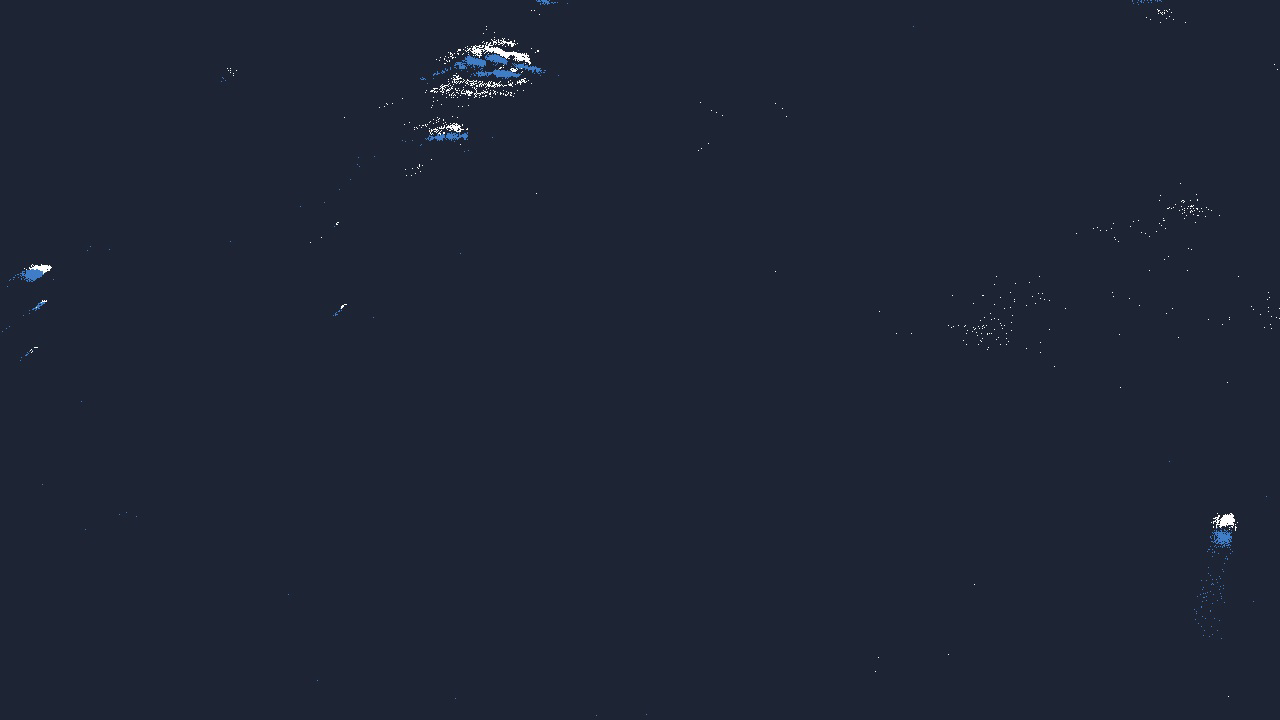}
    \label{fig}
    \end{subfigure}
    \vspace{-8pt}
    \caption{Sample visualized event camera outputs collected during autonomous surface-vehicle operation.}
    \label{fig:eventdata}
\end{figure*}
\vspace{-3pt}
\subsection{Water-Quality Measurements and Complementary Data}
Water-quality measurements from the multiparameter sonde, together with complementary data from the sonar and event camera systems, were obtained during the aforementioned field deployments. Sonde measurements were organized according to sampling site, \ac{GPS} location, depth, and acquisition time, allowing the collected parameters to be compared spatially and temporally. These measurements provide the primary environmental data used to characterize conditions associated with algal bloom development, while the sonar and event-camera recordings provide additional contextual information about the surrounding environment during autonomous operation.

At each sampling location, the sonde recorded chlorophyll-a fluorescence, \ac{FDOM} fluorescence, turbidity, conductivity, salinity, resistivity, density, total dissolved solids, BGA-PC fluorescence, nitrate concentration, temperature, barometric pressure, pressure, depth, and voltage. \figurename~\ref{fig:data_collected} illustrates an example of BGA-PC fluorescence recorded at one of the sampling sites.

In addition to the water-quality measurements, the sonar and event camera systems collected complementary data during vehicle operation. The \acf{SS} sonar provided acoustic imagery of the lake bed structure, while the event camera recorded visual changes on the water surface. Visualized outputs from these sensors are shown in \figurename~\ref{fig:sidescan} and \figurename~\ref{fig:eventdata}. This demonstrates the ability of the platform to collect multi-modal environmental information alongside the primary water-quality measurements.




\vspace{-3pt}
\subsection{Algal Bloom Classification}
\begin{figure}
    \centering
    \includegraphics[width=0.8\linewidth]{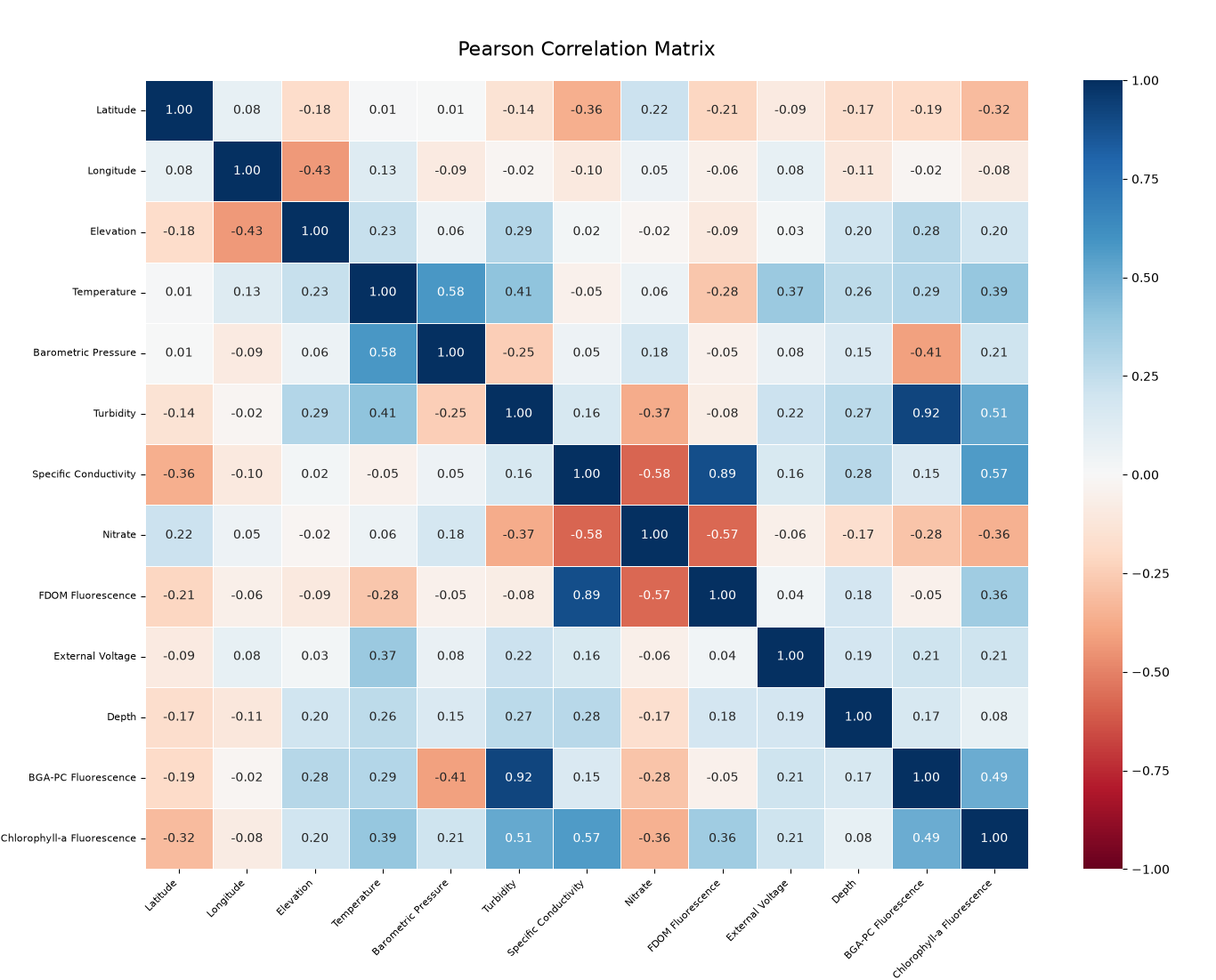}
    \caption{Pearson correlation matrix of input features used to train the final machine learning model}
    \label{fig:pearson}
    \vspace{-15pt}
\end{figure}

\begin{figure}
    \centering
    \includegraphics[width=0.8\linewidth]{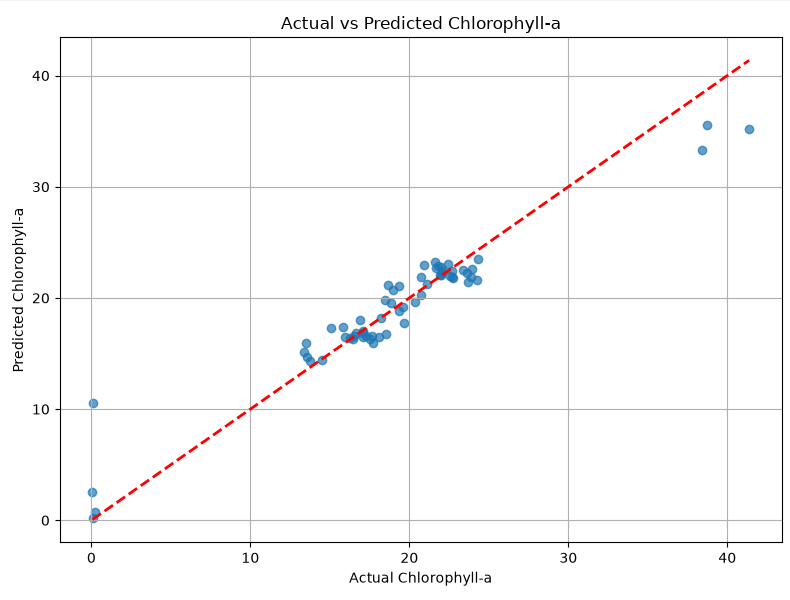}
    \caption{Actual vs Predicted chlorophyll-a concentrations produced by machine learning model}
    \label{fig:algo_performance}
    \vspace{-15pt}
\end{figure}

Once measurements were collected, all possible input features were passed into a Pearson correlation matrix, shown in \figurename~\ref{fig:pearson}, to remove variables with high correlations, as outlined in \cite{ML-Extrusion}. After the correlated variables were removed, the first layer of the model was trained using the Random Forest Predictor model on the given input features to predict chlorophyll-a concentration. The model was trained using 323 data points, 250 trials, and an eighty-twenty training testing split. As illustrated in \figurename~\ref{fig:algo_performance}, the model had a prediction accuracy of 90\%, a testing $R^2$  value of 0.9952, a testing RMSE value of 0.4946, and a testing MAE value of 0.1498.  
The second layer of the model then classifies the predicted chlorophyll-a concentration as \ac{HAB} or non-\ac{HAB}. Consistent with the threshold used in \cite{ALSINI2025103260}, any predicted chlorophyll-a value greater than 10 µg/L was labeled as a \ac{HAB}. Despite the precision of the first layer, the classification layer achieved an overall precision of 99.6\%. 

\vspace{-10pt}
\section{Conclusion}
\label{sec:conclusions}
This paper presented the design and field deployment of an \ac{ASV}-based system for in-situ water-quality measurements related to \acf{HAB} monitoring. The platform integrated autonomous waypoint navigation, depth-controlled sonde deployment, and complementary sensing from sonar and event-camera systems to support spatially distributed environmental data collection. Field deployments demonstrated the ability of the system to collect water-quality measurements across multiple sampling locations and depths with limited manual intervention. Overall, the results highlight the potential of \acp{ASV} as practical and scalable platforms for future \ac{HAB} monitoring applications. Future work will focus on expanded field deployments, additional validation of the classification approach, and further integration of the complementary sensing modalities into the autonomous monitoring workflow.



\section*{ACKNOWLEDGMENT}
\small{ This study was supported by the U.S. Department of the Treasury through the Louisiana Coastal
Protection and Restoration Authority’s Center of Excellence Research Grants Program under the
Resources and Ecosystems Sustainability, Tourist Opportunities, and Revised Economies of the Gulf
Coast States Act of 2012 (RESTORE Act) (Award No. 1 RCEGR260007-01-01). The statements, findings,
conclusions, and recommendations are those of the authors and do not necessarily reflect the views of the
Department of the Treasury.
Co-author Boldor acknowledges the partial support from USDA NIFA Hatch Program (Project \#LAB94672). Published with the approval of the Director of the Louisiana Agricultural Experiment Station as manuscript \# 2026-232-41049.
}

\vspace{-5pt}

\bibliography{library} 
\bibliographystyle{IEEEtran}

\end{document}